# Automated Web-Based Malaria Detection System with Machine Learning and Deep Learning Techniques

**Abraham G Taye[1,2] · Eshetu Negash[1] · Moges Abebe[1] · Sador Yonas[1] · Yared Minwyelet[1]· Melkamu Hunegnaw Asmare[1,3]**

[1] Center of Biomedical Engineering, Addis Ababa Institute of Technology, Addis Ababa University, King George VI St Addis Ababa 1000 Addis Ababa, Ethiopia.

[2] Opus College of Engineering, Marquette University, 1250 W Wisconsin Ave, Milwaukee, WI, 53233

[3] Leuven Center for Affordable Healthcare Technology, KU Leuven| Campus Groep T, Andreas Vesaliusstraat 13, 3000 Leuven, BELGIUM

**Abstract:** Malaria parasites pose a significant global health burden, causing widespread suffering and mortality. Detecting malaria infection accurately is crucial for effective treatment and control. However, existing automated detection techniques have shown limitations in terms of accuracy and generalizability. Many studies have focused on specific features without exploring more comprehensive approaches. In our case, we formulate a deep learning technique for malaria-infected cell classification using traditional CNNs and transfer learning models notably VGG19, InceptionV3, and Xception. The models were trained using NIH datasets and tested using different performance metrics such as accuracy, precision, re- call, and F1-score. The test results showed that deep CNNs achieved the highest accuracy - 97%, followed by Xception with an accuracy of 95%. A machine learning model SVM achieved an accuracy of 83%, while an Inception-V3 achieved an accuracy of 94%. Furthermore, the system can be accessed through a web interface, where users can upload blood smear images for malaria detection.

**Key Words:** Malaria, Automated Detection, Machine Learning, Deep Convolutional Neural Networks, Transfer Learning

## 1. Introduction

### Background

Malaria is a life-threatening infectious disease, that has a significant impact on global health. As of more recent data, it is estimated that around half of the world's population is at risk of contracting malaria. In 2019, there were approximately 229 million cases of malaria reported worldwide. This resulted in an estimated 409,000 deaths, affecting vulnerable populations such as pregnant women and infants and small children five years old. Malaria continues to pose a substantial burden on public health, emphasizing the urgent need for effective prevention, diagnosis, and treatment strategies [1].

Malaria is a deadly human disease caused by organisms called Plasmodium genus. Malaria infections are largely spread among humans via the bites of so-called malaria vectors — adult females belonging to Anopheles-type mosquitos. Widely speaking, Plasmodium infection is caused by many species of Plasmodium genus protozoa [2]. Figure one shows the lifecycle of the Plasmodium parasite. All five species' life and infective cycles are similar. And their morphology and body shapes look alike on animals [3].

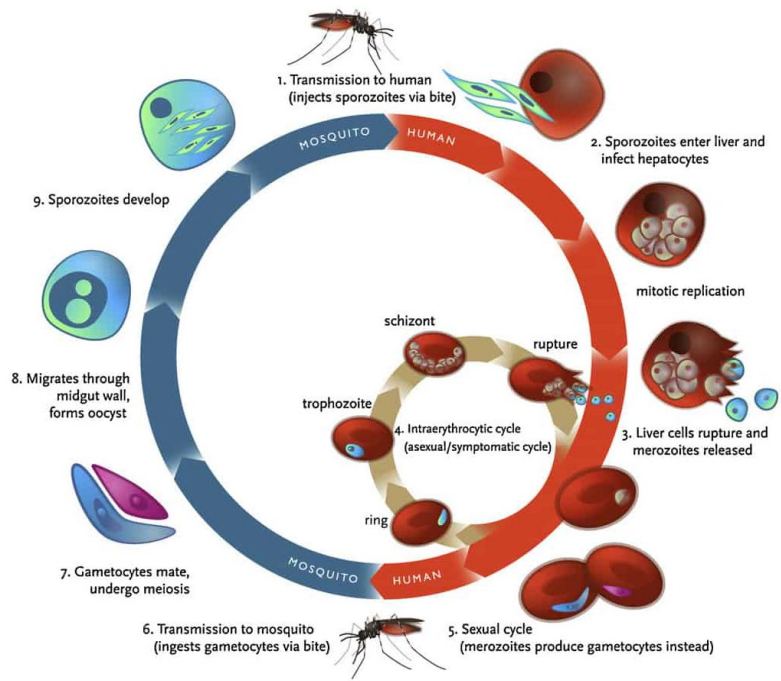

*Figure 1: The life cycle of malaria parasite [5*

According to the World Health Organization (WHO) statistics, malaria infection causes over one million human infections each year, with a particularly alarming impact in certain regions. For instance, as referenced in [5], there were approximately 219 million reported cases of malaria across 87 countries affected by the epidemic. Africa is the most affected continent with 95% of all malaria cases reported and 96% of all deaths [4,5]. The WHO report also highlights the devastating toll of malaria in Africa, where it accounts for a significant proportion of the top ten causes of mortality.

Malaria is among the most prominent reasons for being ill or dying. Malaria infections have been recorded in more than 75 percent of the landscapes below 2000m (about 1.24 mi) elevated levels in Ethiopia [7]. In Ethiopia, the main epidemics happen every five–eight years while focal ones are happening annually [8]. The population of Ethiopia is more than 100 million and over 68 percent of this population is at risk [7, 9]. Annually, there are approximately 2.9 million of the malaria patients and this leads to malaria death, increasing sharply in epidemics while morbidity and mortality dramatically increase.

There is a need for early detection and quick care and management of malaria. It is a paramount importance in reducing death rates linked with the disease. It is recommended to initiate treatment promptly, preferably 24 hours after having had a fever, particularly in vulnerable populations such as children under five years of age [10]. Early intervention is fundamental in preventing severe complications and reducing mortality rates.

In Sub-Saharan countries, one of the major priorities is the installation of early-warning mechanisms for epidemics against malaria. There is a need for novel diagnostic approaches because current methods used in diagnostics like microscopic vision analysis of blood smears and RDTs have their drawbacks. An exciting breakthrough in this area is the introduction of image analysis methods using deep learning which is a sub-domain of Artificial Intelligence. This automated approach holds great potential to complement and enhance current diagnostic tools for more accurate and efficient malaria diagnosis.

Computer-aided diagnosis using deep learning techniques has shown promising results in medical imaging analysis, including malaria classification. In this context, a promising new avenue emerges with the application of supervised machine learning techniques on health sector. By leveraging machine learning algorithms in malaria classification systems, we can achieve a significant improvement in the automation of medical image analysis. By implementing an automated malaria detection system that utilizes images extracted from blood smear films, we can ensure more precise measurements in malaria diagnosis. This, in turn, would lead to reduced delays in treating patients and alleviate the burden on physicians who currently spend considerable time on diagnosis.

Bearing the above-mentioned malaria diagnostic techniques, automated detection systems are being developed by different researchers to enhance and combat the prevalence of the disease. The application of supervised machine learning techniques has revolutionized medical image analysis, particularly in the field of malaria detection systems.

This paper is motivated by the objective of implementing an automatic malaria detection system that utilizes supervised machine learning methods to achieve precise diagnosis and efficient detection of the malaria parasite. We analyzed a deep learning approach for malaria classification using traditional CNNs and transfer learning models such as VGG19, InceptionV3, and Xception.

Therefore, the motivation behind this paper lies in the imperative to develop an automated malaria classification system that empowers accurate and precise diagnosis by employing supervised machine learning approaches. By harnessing the power of these advanced techniques, we aim to enhance the efficiency of malaria parasite detection, contributing to improved healthcare outcomes for individuals affected by this devastating disease.

## 2. Literature Review

The existing literature on malaria detection encompasses a wide range of studies conducted in both endemic and non-endemic regions. Researchers have explored various aspects of malaria detection, including the identification and classification of malaria parasites, feature extraction techniques, machine learning algorithms, and the integration of innovative diagnostic tools. These studies have yielded valuable insights into the strengths, limitations, and potential applications of different detection approaches.

### 2.1. Literatures on Automated Malaria Diagnosis

In this section, we provide comprehensive information about our findings on automated microscopy for diagnosing malaria. We have compiled a vast number of references, covering many articles published on this topic, especially those from the past decade.

The field of automated cell microscopy for malaria diagnosis encompasses a wide range of research. Generally, most systems adhere to a set of essential processes. The subsequent step usually revolves around detecting and segmenting individual blood cells, such as infected or uninfected. The RBC detection and segmentation section provides various segmentation methods utilized for diagnosing malaria microscopically. After cell segmentation, many articles have proposed methods for characterizing an attribute vector summarizing the appearance of the segments. Different features, as well as strategies of feature selection presented in the literature are presented in the Feature Extraction and Selection section.

In the field of malaria detection, researchers have employed diverse techniques and methodologies, ranging from traditional microscopic examination of blood smears to advanced computer-based analysis using machine learning and image processing algorithms. These algorithms have shown promise in improving the speed and efficiency of conducting malaria diagnosis.

### 2.2. Deep Learning and Transfer Learning for Malaria Detection

The paper "Deep Learning and Transfer Learning for Malaria Detection" by Singh et al advocates for automated diagnostic processes using deep learning technologies. The objective of the paper is to enhance diagnostic accuracy by evaluating parasitemia in microscopic blood slides. The paper discusses using transfer learning and fine-tuning methods for training CNNs such as ResNet50, ResNet34, and VGG-19 on a dataset of stained pictures of infected and uninfected erythrocyte. This approach uses the variance principle as it characterizes both the intensity of the Plasmodium parasites and the RBCs. On this note, the study reveals that the VGG-19 model had the highest accuracy scores, based on the parameters considered and employed data set. This study is consistent with previous findings that show that deep learning algorithms can detect malaria. The results have been applied in constructing more advanced deep learning-based malaria disease diagnosis device. [35]

### 2.2.1. Malarial Parasite Classification using Recurrent Neural Network

The paper proposes use of a recurrent neural network (RNN) for classification of segmented red blood cells (RBCs) into normal RBCs and infected cells. The infected cells are further classified into falciparum and vivax plasmodium. The paper acknowledges that the complexity of cell nature and image uncertainties make cell segmentation and morphological analysis among the challenging problems for malaria. Proposed method has been developed to increase the accuracy of segmenting the parasites and classifying them as malaria or not. [36]

### 2.2.2. ImageNet Classification with Deep Convolutional Neural Networks

The paper titled "ImageNet Classification with Deep Convolutional Neural Networks" by Alex Krizhevsky, Ilya Sutskever, and Geoffrey Hinton obtained the top-1 and top-5 percentages of 37.5% and 17.0% respectively, with this being superior to the best reported results until that time. It was made up of five Convolutional layers with most of their preceding max-pooling layers followed by three final fully connected layers which ended in a 1K-way SoftMax Authors used non-saturated neurons, and the implemented convolution operation on the GPU allowed to make the training faster. Additionally, they implemented dropout – a modern, and regularization method for minimized overfitting of the fully connected layers. Nevertheless, the variants of these models were submitted to the ILSVRC-2012 competition and achieved the first five test errors of 15.3%. [37]

### 2.2.3. Towards Low-Cost and Efficient Malaria Detection

This paper Towards Low Cost and Efficient Malaria Detection contributes to additional research into malaria microscopy using low magnifications at affordable cost. The authors highlight the importance of early and correct diagnosis of malaria to avoid health complications, which is currently dependent access to expensive microscopes and experienced professionals in reading smears. The model implies that we can assist expert's load by improving cheap microscopes diagnostic accuracy using deep learning-derived methods. However, the absence of a reasonable-size dataset is a challenge in this area.

However, in order to solve this problem, the article presents a large database of images containing photographs of smears in the blood of various malaria-infected patients examined using microscopes from distinct levels of cost and in multiple degrees of magnification. An extension of a partially supervised domain adapter is also proposed to apply the object detector in images obtained using the low-cost microscope. In essence, the work serves a good cause in developing cost-effective and efficient malaria detecting techniques with the aid of a large-scale data set as well as deep learning concepts. [38]

### 2.2.4. Intelligent Deep Transfer Learning Based Malaria Parasite Detection and Classification Model Using Biomedical Image

An intelligent deep transfer learning-based malaria parasite detection and classification (IDTL-MPDC) model using Biomedical images in this paper. Firstly, this approach entails performing a digital filtering method called median filtering on the input image to remove digital noise and thereafter the denoising process is carried out. The Res2Net model with its best-tuned input parameters implemented through DE method is used as a classifier for extracting the relevant feature vectors. KNN classifier is employed to determine the appropriate classes and the blood smear images are classified. A benchmark data set is used for evaluating the performance of the IDTL–MPDC technique whose accuracy is 95.86%, sensitivity of 95.82%, specificity of 95.98%, and F1 score of 95.69% [41]

In this study, we discussed various ways of applying deep-learning techniques in medical imaging, looked at Convolutional Neural Networks, and studied some aspects of Transfer Learning as its application.

## 3. Methods and Materials

### 3.1. Dataset and Data Preprocessing

In our study, we utilized datasets sourced from the National Institute of Health (NIH) website, specifically accessing a malaria dataset that proved to be a valuable resource for our research [42]. The NIH dataset consisted of a total of 27,557 images, with 13,778 categorized as parasitized (malaria-infected) and 13,779 categorized as healthy (non-infected) samples. To ensure consistency and facilitate our analysis, we initially categorized the image files based on their corresponding tags, distinguishing between healthy and malaria-infected samples.

Figure 3 and 4 shows microscope images of healthy and parasitized blood cells from NIH dataset

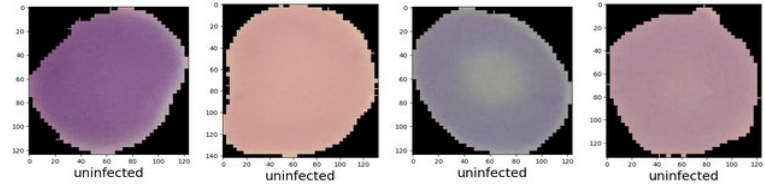

*Figure 2: Uninfected dataset microscope images acquired from NIH*

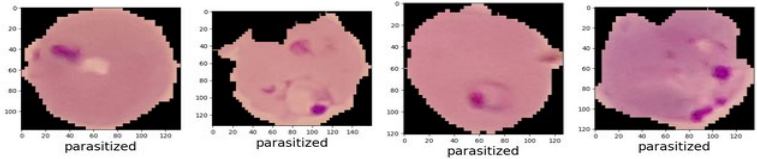

*Figure 3: Parasitized dataset microscope images acquired from NIH*

In addition to the challenges of limited availability, the local datasets we collected posed another issue. These datasets had low resolution due to the utilization of low-cost microscopes, resulting in blurry images that lacked the necessary level of detail for accurate segmentation and training of our model. In contrast, the NIH datasets were extracted from high-cost microscopes, providing more detailed features that were essential for training our model effectively. The disparity in image quality between the local datasets and the NIH dataset further emphasized the need for us to rely on the NIH dataset as a valuable resource for our research on malaria detection using transfer learning.

### 3.2. Pre-Processing of Datasets

Preprocessing includes transforming raw data into a format that can be easily used by the model. In this case, the pre-processing includes resizing and rescaling the images in the dataset to improve performance and reduce computation time. By performing data augmentation, normalization, and splitting, the models can learn meaningful features from the data and achieve better performance in image classification tasks. [45]

- Data augmentation: Different types of transformation were performed on the training images such as horizontal and vertical flipping, rotation, shear and movement.
- Data normalization: Standardization of the pixel values among the image files was achieved by scaling the pixel values down to a 0-1 range.
- Data splitting: Train, valid, and test datasets were partitioned from that dataset. For the training of the machine learning models, we used the training set, for tuning of hyperparameters and evaluation of the performance of the models – the validation set, and finally to test the final performance of the models – the test set.

- **3.3 Training Methods**

In the development of an automated malaria classification system, various machine learning and deep learning methods can be used to train the models. In this project, we utilized three core training methods, which are machine learning, deep learning, and transfer learning algorithms.

- Machine Learning: We used a Support Vector Machine (SVM) algorithm to train the model. SVM is a supervised learning algorithm that can be used to identify the boundary between the classes and classify the images as either infected or uninfected.
- Deep Learning: Convolutional neural networks (CNNs) are used to train the model that can learn features from the images by applying convolution operations on the pixels.
- Transfer Learning: We used pre-trained models such as VGG19, InceptionV3, and Xception to train the model to classify the cell images as either infected or uninfected.

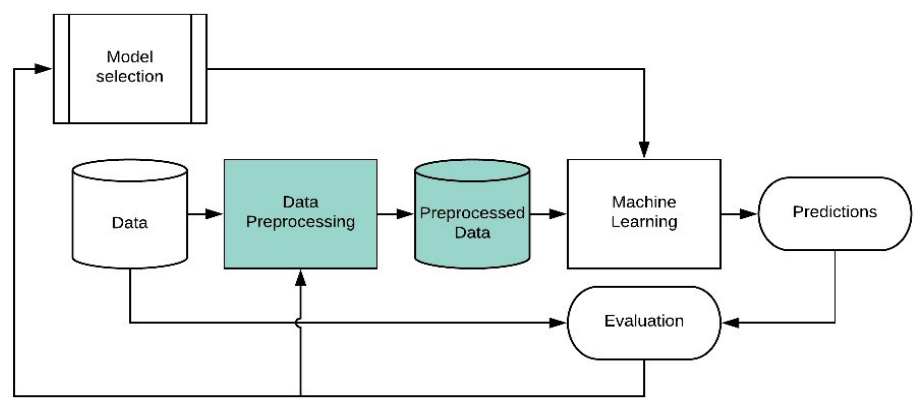

*Figure 4: Workflow of malaria model training*

Figure 5 illustrates a possible approach for malaria disease classification and analysis. Malaria infection images are initially gathered and sorted into different groups. The images are then fed into the proposed technique for training the model in the next stage. The newly trained architectural model is utilized to predict images that are previously unnoticed.

### 3.2.1. Machine Learning Experiments

Machine learning encompasses the development of algorithms and techniques that enable machines to automatically learn and make precise predictions based on past observations. It involves the extraction of information from data through computational and statistical methods. As highlighted by [44], machine learning (ML) is a subfield of computer science that focuses on studying and constructing algorithms capable of learning from data and making predictions.

### 3.2.1.1. Support Vector Machines

In our paper, the emphasis lies on SVM, which is a powerful supervised machine learning algorithm known for its capability to find optimal decision boundaries in high-dimensional spaces, making it well-suited for classification tasks. Support Vector Machines (SVM) is a supervised machine learning algorithm that has gained significant attention and proven to be effective in various domains, including medical image analysis. In the case of malaria detection, where blood smear images can contain numerous features, SVM can effectively handle the complexity and variability of the data. By employing a kernel function.

Nonetheless, SVM can deal with linearly separable and non-linearly separable data. The purpose of SVM is to find the best dividing line (hyperplane) in the data space (feature space), which will distinguish one class from another as much as possible. This characteristic allows SVM to handle complex and overlapping class distributions, which can be crucial in distinguishing between infected and uninfected blood smear images in malaria diagnosis.[45]

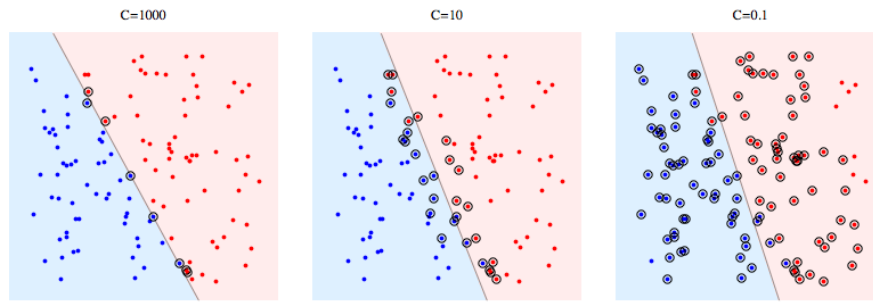

*Figure 5: Comparing the Impact of Penalty Parameter C on Decision Boundary and Misclassification Term at Different Values (C=100, C=10, C=1) [48]*

This image demonstrates the importance of hyper parameterization in optimizing SVM performance. The penalty parameter C represents the misclassification or error term, which tells the SVM optimization how much error is bearable.

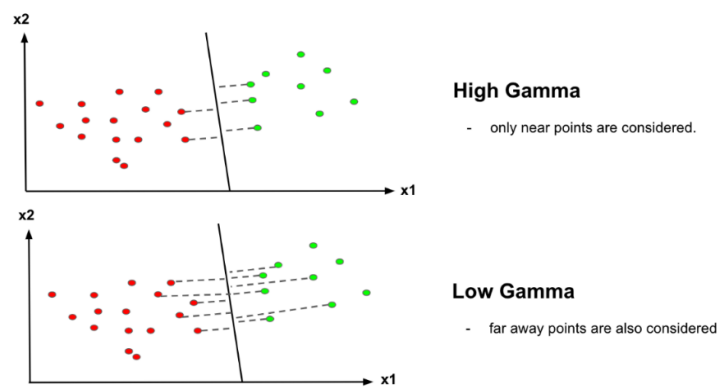

*Figure 6: Fine-Tuning SVM Performance through Gamma Parameter Optimization [49]*

This image highlights the importance of the gamma parameter in SVM optimization. Gamma determines how far the calculation of plausible line of separation extends. When gamma is higher, nearby points have a greater influence on the decision boundary, resulting in a more complex and tightly fitted model. Conversely, lower gamma values result in a smoother decision boundary that considers a wider range of points, including those that are further away.

### 3.2.2.    Deep Learning Experiments

### 3.2.2.1.  Convolutional Neural Network

We designed a traditional CNN with three convolutional layers followed by max-pooling layers and two fully connected layers.

CNNs have been shown to be highly effective for a wide range of image classification tasks, including malaria detection. They can learn complex features and patterns in the images, and can generalize well to new, unseen samples.

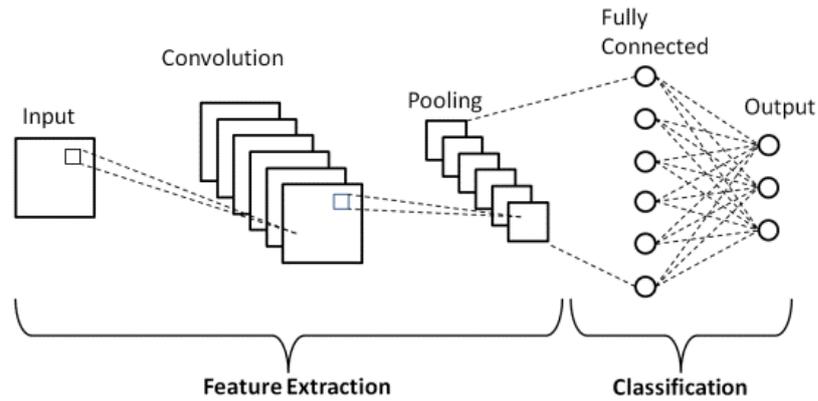

*Figure 7: Outline of Traditional CNN Architecture [49]*

In the case of malaria detection, CNN is trained on a dataset of labeled medical images, with the goal of learning to accurately distinguish between images with and without malaria.

**A. Deep CNN with dropout, weight initialization:**

We used a deep convolutional neural network (CNN) model with dropout, batch normalization, and RMSprop optimizer to perform image classification on a dataset of images. The model was implemented using the Keras Sequential API in Python. The model architecture consisted of a series of convolutional layers followed by max pooling layers, with dropout layers to reduce overfitting.

- The specific architecture of the model included a series of three stacked convolutional layers comprising increasing number of filters, which are then followed by batch normalization and a layer of dropout.
- It consisted of converting convolutional layers outputs to a vector and feeding this vector to the 256-unit dense layer with ReLU activation and batch normal and dropout after.
- Denser output layer of 2 units with the softmax activation function yielding probability distribution across two the classes.

Overall, the use of a deep CNN with dropout and batch normalization can improve the performance of image classification models by reducing overfitting and improving the ability of the network to learn complex features from the input images.

**B. Deep CNN with dropout, weight initialization, and zero padding:**

Our study utilized a deep convolutional neural network (CNN) model with dropout, weight initialization, and zero padding to perform image classification on a dataset of images. The model architecture consisted of a series of convolutional layers followed by max pooling layers, with dropout layers to reduce overfitting. The specific architecture of the model was as follows:

- The first layer was a 2D convolutional layer with 32 filters of size 3x3 and the ReLU, with an input shape of (224, 224, 3). This was followed by a zero-padding layer
- Next, there were two more 2D convolutional layers with 32 filters of size 3x3 and the ReLU, followed by a max pooling layer with a pool size of 2x2, and a dropout layer
- The next three sets of layers followed a similar pattern, with 64 filters of size 3x3 in the convolutional layers, and a max pooling layer with a pool size of 2x2, and a dropout layer

- The final convolutional layer had 128 filters of size 3x3, followed by another max pooling layer with a pool size of 2x2, and a dropout layer with a rate of dropout_conv.
- The output of the convolutional layers was flattened and fed into a dense layer with 256 units and the ReLU activation function, followed by another dropout layer.
- The final output layer was a dense layer with 2 units and the softmax activation function, which output a probability distribution over the two classes.

The model was evaluated on the testing set using accuracy, precision, recall, and F1-score metrics, and the results were compared to a baseline model to assess the efficiency.

### 3.2.3. Transfer Learning

Traditional machine learning utilizes training and testing datasets with identical data distributions and feature spaces. The optimization and training of the model is a difficult and time-consuming process. The training requires a strong graphics processing unit (GPU) as well as millions of training dataset. [45] The transfer-learning is one of such powerful techniques because it utilizes knowledge acquired during the training of the pre-trained model for its initialization phase, and further adjusts itself by incorporating new information.

In this study, we will be exploring the use of transfer learning for malaria detection, using different pre-trained CNN models such as VGG, Xception, and InceptionV3. These models have been trained on large image datasets and have learned a set of general features that are transferable to other image classification tasks, including malaria detection.

Our transfer learning approach involves training three different model architectures to demonstrate the impact of incremental unfreezing and fine-tuning on model accuracy. The three models are as follows:

i.   *Model Architecture* 1: Pre-trained model as a feature extractor with frozen layers
ii.  *Model Architecture* 2: Pre-trained model as a feature extractor with incremental unfreezing and fine-tuning
iii. *Model Architecture* 3: Pre-trained model as a feature extractor with entire network unfreezing and fine-tuning

**Model Architecture 1:**

In the first model architecture, we used the pre-trained ImageNet model such as InceptionV3, VGG19, and Xception as a feature extractor and added fully connected layers on top to classify images. We froze the weights of the pre-trained model and trained only the top layers using categorical cross-entropy loss and the stochastic gradient descent (SGD) optimizer.

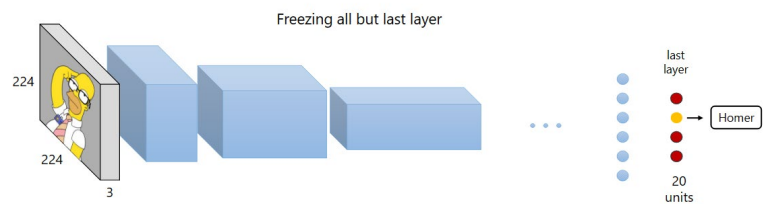

*Fig: Model Architecure 1: Freezing all but last layer*

- Used pre-trained model as feature extractor.
- Added fully connected layers on top for classification
- Froze pre-trained model weights and trained only top layers
- Used categorical cross-entropy loss and SGD optimizer
- Trained on training set, evaluated on validation set with accuracy and loss
- Implemented callbacks including early stopping, model, and learning rate reduction on plateau.

**Model Architecture 2:**

In the second model architecture, we used the pre-trained ImageNet model as a feature extractor and incrementally unfroze layers to fine-tune the entire network. We first set all layers except the last two blocks to non-trainable and trained the top layers using categorical cross-entropy loss and the SGD optimizer. Then, we unfroze the last two blocks and fine-tuned the entire network using the same loss function and optimizer.

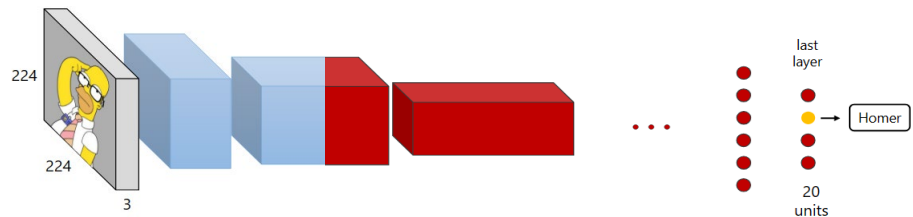

- Used pre-trained model as feature extractor
- Incrementally unfroze layers to fine-tune entire network
- Trained top layers with non-trainable layers (except last 2 blocks)
- Unfroze last 2 blocks and fine-tuned entire network
- Used categorical cross-entropy loss, SGD optimizer,
- Implemented callbacks including early stopping, model checkpointing, and learning rate reduction on plateau
- Preprocessed images by resizing to 128x128 and normalizing pixel values
- Trained on training set, evaluated on validation set with accuracy and loss metrics.

**Model Architecture 3:**

In the third model architecture, we fine-tuned the entire pretrained ImageNet model from scratch for our binary classification task. We initialized the weights of the model with pre-trained weights and trained the entire network using categorical cross-entropy loss and the SGD optimizer. We also implemented the same callbacks.

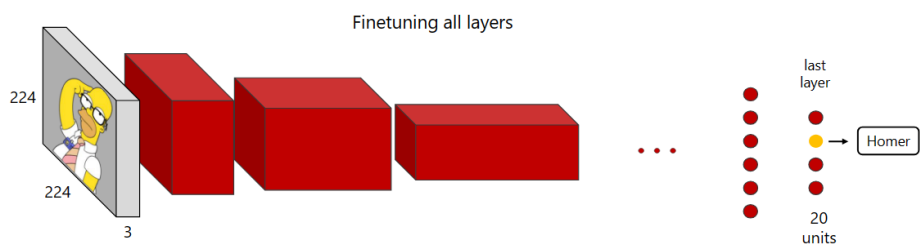

*Fig: Model Architecture 3: Fine tuning all layers*

- Fine-tuned entire network from scratch with pre-trained weights
- Used categorical cross-entropy loss, SGD optimizer, and implemented callbacks including early stopping, model checkpointing, and learning rate reduction on plateau.
- Preprocessed images by resizing to 128x128 and normalizing pixel values
- Trained on training set, evaluated on validation set with accuracy and loss metrics.

### 3.2.3.1. VGG-19

VGG-19 is a 19-layer deep Convolutional Neural Network is made up of 16 convolutional layers, 3 fully linked layers, and five max-pool layers. It is an image database with 14,197,122 pictures arranged in the WordNet hierarchy.[45] We used the pre-trained VGG19 model, which consists of 19 layers, including 16 convolutional layers and three fully connected layers.

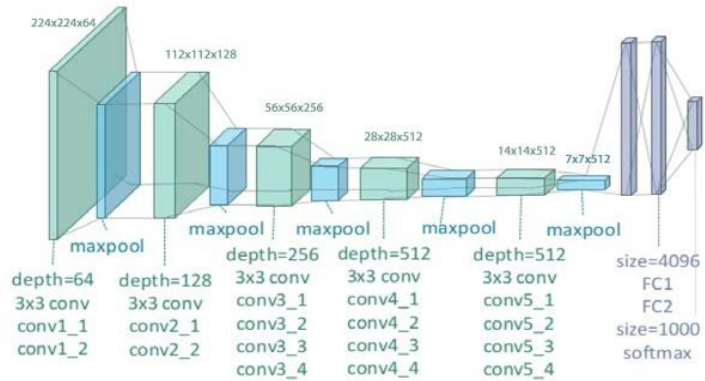

*Figure 8: Illustration of the network architecture of VGG-19 model [51]*

### 3.2.3.2. Inception-V3

Inception-v3 is a convolutional neural network architecture from the Inception family that has been shown to attain greater than 78.1% accuracy on the ImageNet dataset1.[45] The Inception architecture introduces various inception blocks, which contain multiple convolutional and pooling layers stacked together, to give better results and reduce computation costs.

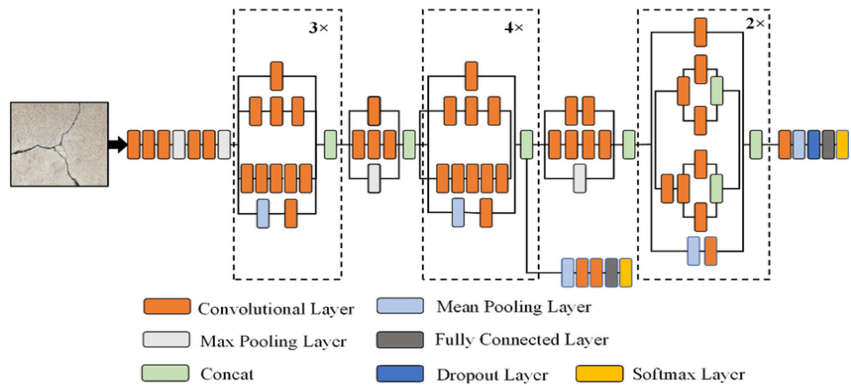

*Figure 9: Inception-V3 model architecture [52]*

### 3.2.3.3. Xception

Xception is a deep convolutional neural network architecture that has shown excellent performance on the ImageNet dataset. It is based on the idea of depth-wise separable convolutions, which can reduce the number of parameters and computation while maintaining high accuracy.

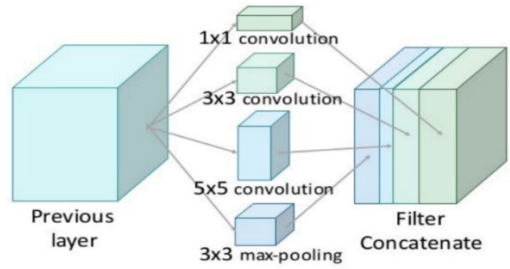

*Figure 10: Xception: Deep Learning with Depth-wise Separable Convolutions [53]*

### 3.3. Web Demo using TensorFlow.js.

In this section, we discuss the architecture and implementation of the web demo using TensorFlow.js for malaria detection. The web demo allows users to upload an image of a blood smear and receive a prediction of whether it is infected with malaria or not.[45]

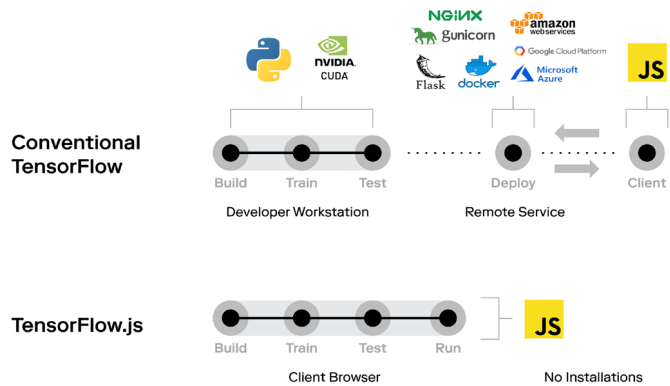

*Figure 11: The Evolution of Machine Learning Architecture: Conventional ML vs. TensorFlow.js [54]*

The figure depicts the conventional ML architecture with a diverse tech stack that requires coordination and communication between multiple services, making it challenging to deploy machine learning models on websites. In contrast, TensorFlow.js has the potential to revolutionize the future of machine learning by offering a streamlined approach to deploying and executing machine learning models directly within web applications, enabling developers to build intelligent applications that can process data and make predictions in real-time. [54]

### 3.3.1. Deployment and Testing

In this section, we discuss the deployment and testing of the web demo using TensorFlow.js for malaria detection. The web demo is deployed on a cloud-based platform, which allows for easy scalability and accessibility.

The architecture and implementation of the web demo are based on a front-end and back-end model, with the back end using TensorFlow.js to perform the prediction. The deployment and testing of the web demo involves several steps, including continuous integration and deployment and various types of testing.

**4.1 Result and Discussion**

**4.2 Performance Metrics**

In this section, we discuss the metrics used for performance evaluation of malaria detection. The selection of appropriate performance metrics is crucial in assessing the effectiveness of the models in real-world applications.

- **Accuracy:** The accuracy is the fraction of correctly classified samples over the total number of samples. It is computed as follow

$$\text{Accuracy} = \frac{(True\ Positive + True\ Negative)}{(True\ Positive + True\ Negative + False\ Positive + False\ Negative)}$$

- **Precision:** The precision is the fraction of true positives (TP) over the sum of true positives and false positives (FP). It represents the proportion of positive predictions that are correct. It is computed as follows:

$$\text{Precision} = \frac{(TP)}{(TP + FP)}$$

- **Recall (Sensitivity):** The recall (or sensitivity) is the fraction of true positives over the sum of true positives and false negatives (FN). It is computed as follows:

$$\text{Recall} = \frac{(TP)}{(TP + FN)}$$

- **Specificity:** The specificity is the fraction of true negatives (TN) over the sum of true negatives and false positives (FP). Specificity represents the proportion of actual negatives that are correctly identified by the model. It is computed as follows:

$$\text{Specificity} = \frac{(TN)}{(TN + FP)}$$

- **F1-score:** The F1-score is the harmonic mean of precision and recall, defined as 2 * (precision * recall) / (precision + recall). It represents the balance between the precision and recall.

$$\text{F1} - \text{score} = 2 \times \frac{(Precision\ \times Recall\ )}{(Precision + Recal)}$$

- **Sensitivity (Recall):** Sensitivity is another name for recall, which is the division of true positives over the sum of TP and FN. It represents the proportion of actual positives that are correctly identified by the model.

- **Specificity:** Specificity is the fraction of true negatives over the sum of true negatives and false positives. It represents the proportion of actual negatives that are identified

- **Macro Average:** Macro average computes the metric independently for each class and takes the average across all classes.

- **Weighted Average:** Weighted average computes the metric for each class weighted by the number of samples in that class and takes the averages.

- **AUC-ROC** is a plot of true positive rate (TPR) against false positive rate (FPR) at different classification thresholds. It provides a measure of the model's ability to distinguish between positive and negative cases.

## 4.2. Machine Learning Results

### 4.2.1 SVM Results

- **Data Split**

The dataset used for this SVM machine learning model consists of malaria-infected and uninfected cells, totaling 13,779 images. The dataset was randomly split into two sets: 11,023 images for the training set, and 2,756 images for the testing set. In addition to the data split, a typical train-validate-test split was used to further divide the data into three subsets: a training set, a validation set, and a testing set. The training set consisted of 70% of the data (11,023 images), the validation set consisted of 15% of the data (approximately 2,067 images), and the test set consisted of 15% of the data (approximately 2,067 images).

The test set was kept hidden throughout the model development and training process to ensure that the model's performance could be accurately evaluated on unseen data. The data split was done randomly to ensure that each subset was representative of the entire dataset. The images were randomly shuffled and then split into the desired subsets to avoid any biases that might arise if the data were split in a non-random way. Overall, an SVM model was developed for malaria classification that achieved high accuracy on the testing set.

- **Model Results**

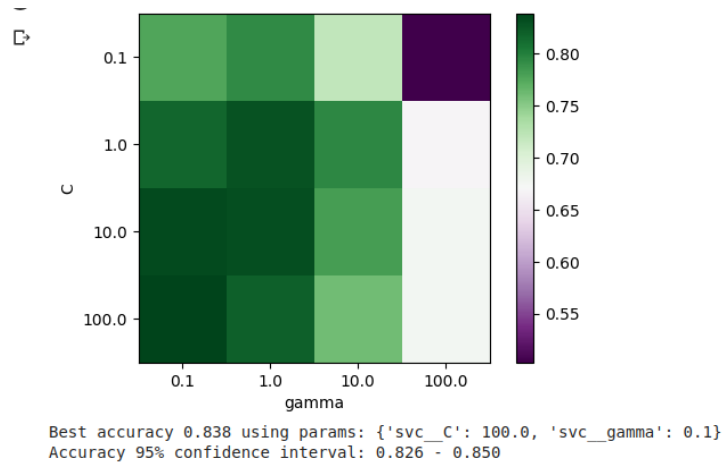

Best accuracy 0.838 using params: {'svc__C': 100.0, 'svc__gamma': 0.1}
Accuracy 95% confidence interval: 0.826 - 0.850

*Figure 12: Optimizing SVM Performance: Grid Search Results and Model Training*

The hyperparameters for the SVM model were tuned using a exhaustive search over the hyperparameters in the specified ranges of values. The C and gamma hyperparameters for the RBF kernel were tuned over a range of values to identify a good starting point for the grid search. The grid search was performed over a wider range of hyperparameters using cross-validation to estimate the performance of each combination of hyperparameters. The range of hyperparameters searched over were C values from 0.1 to 100 and gamma values from 0.001 to 1.

The optimal hyperparameters for the SVM model were determined to be C=1 and gamma=0.01. These hyperparameters were used to train the final SVM model, which achieved an accuracy of 84% on the testing dataset.

*Table 1: Classification report of SVM model*

|  | Precision | Recall | F1-score | Support |
|---|---|---|---|---|
| **uninfected** | 0.82 | 0.85 | 0.83 | 2756 |
| **infected** | 0.84 | 0.82 | 0.83 | 2756 |
|  |  |  |  |  |
| **Accuracy** |  |  | 0.83 | 5512 |
| **Macro avg** | 0.83 | 0.83 | 0.83 | 5512 |
| **Weighted avg** | 0.83 | 0.83 | 0.83 | 5512 |

The SVM model achieved an accuracy of 83% on the testing dataset, with an average precision of 83%, recall of 83.5%, and F1-score of 83%. The Matthews correlation coefficient (MCC) was calculated to be 0.665, indicating a moderate level of agreement between the predicted and actual labels.

These results suggest that the SVM model performed reasonably well on the classification task, with a good balance between precision and recall. The values obtained for these metrics suggest that the model can perform well in both areas, with a slightly higher specificity than sensitivity. it suggests that the model is better at correctly identifying negatives than positives. In other words, the model may be more conservative in its predictions and more likely to classify a sample as negative to avoid false positives. This may be desirable in certain use cases, such as medical diagnosis where false positives can lead to unnecessary treatment or testing. sensitivity.

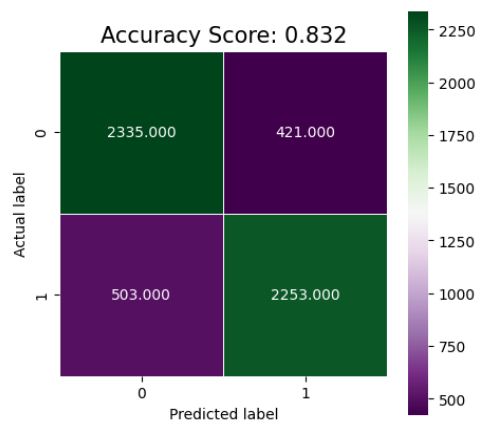

Figure 13: Confusion Matrix for of SVM model

The confusion matrix for the model showed that it correctly classified 2,253 infected cells and 2,335 uninfected cells but misclassified 421 uninfected cells and 503 infected cells. This corresponds to a false positive rate of 15% and a false negative rate of 18%.

### 4.3.    Deep Learning Results

### 4.3.1.    Deep CNN with Drop-out and Weight Initialization

* **Data split**

The dataset used for this classification task consisted of images of malaria-infected and uninfected cells, totaling 27,358 samples. The dataset was split into three subsets using a train-validate-test split, with 23,254 samples randomly selected for the training set, while the remaining 4,104 samples were further split equally into the validation and testing sets. The purpose of this split was to train the model on the training set, tune the model's hyperparameters on the validation set, and evaluate the model's performance on the testing set. The split was done randomly to ensure that each subset was representative of the entire dataset, and the images were shuffled before being split to avoid any biases.

* **Model Result**

A deep CNN model was developed with dropout, weight initialization, and batch normalization for the classification task. The model achieved an accuracy of 96% on the testing set, with an average precision, recall, and F1-score of 96%. The high accuracy, precision, and recall of the model suggest that it was able to effectively classify both infected and uninfected cells.

*Table 2: Classification report of deep CNN model-1*

|  | Precision | Recall | F1-score | Support |
|---|---|---|---|---|
| **uninfected** | 0.94 | 0.97 | 0.96 | 2053 |
| **parasitized** | 0.97 | 0.94 | 0.96 | 2051 |
|  |  |  |  |  |
| **Accuracy** |  |  | 0.96 | 4104 |
| **Macro avg** | 0.96 | 0.96 | 0.96 | 4104 |
| **Weighted avg** | 0.96 | 0.96 | 0.96 | 4104 |

A confusion matrix and classification report were provided to evaluate the model's performance in more detail. The confusion matrix showed that the model correctly classified 3,925 samples, including 1,994 uninfected cells and 1,931 parasitized cells. The model misclassified only 59 uninfected cells and 120 parasitized cells, resulting in a false positive rate of 3% and a false negative rate of 6%. The precision, recall, and F1-score were all high for both classes, indicating that the model had a balanced trade-off between false positives and false negatives.

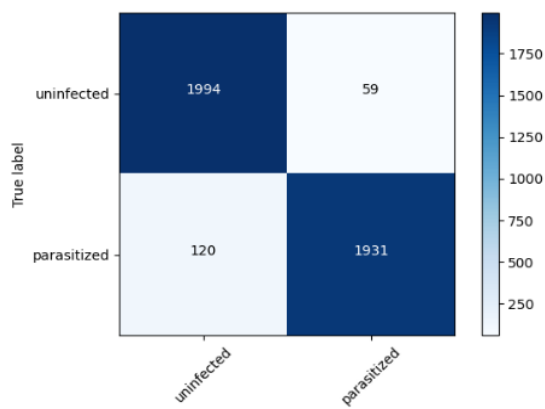

*Figure 14: Confusion Matrix of deep CNN model -1*

### 4.3.1. Deep CNN with drop out, weight initialization and Zero Padding

- **Data split**

The dataset used for this classification task consisted of images of malaria-infected and uninfected cells, totaling 27,358 samples. The dataset was split into three subsets using a train-validate-test split, with 23,254 samples randomly selected for the training set, while the remaining 4,104 samples were further split equally into the validation and testing sets. The split was done randomly to ensure that each subset was representative of the entire dataset, and the images were shuffled before being split to avoid any biases.

- **Model Result**

A deep CNN model was developed with dropout, weight initialization, and batch normalization for the classification task. The model achieved an accuracy of 96% on the testing set, with an average precision, recall, and F1-score of 96%. The high accuracy, precision, and recall of the model suggest that it was able to effectively classify both infected and uninfected cells.

*Table 3: Classification report of deep CNN model -2*

|  | Precision | Recall | F1-score | Support |
|---|---|---|---|---|
| **uninfected** | 0.96 | 0.98 | 0.97 | 2053 |
| **parasitized** | 0.98 | 0.96 | 0.97 | 2051 |
| **Accuracy** |  |  | 0.97 | 4104 |
| **Macro avg** | 0.97 | 0.97 | 0.97 | 4104 |
| **Weighted avg** | 0.97 | 0.97 | 0.97 | 4104 |

To evaluate the model's performance in more detail, a confusion matrix and classification report were generated. The confusion matrix showed that the model correctly classified 3,982 samples, including 2,004 uninfected cells and 1,978 parasitized cells. The model misclassified only 49 uninfected cells and 73 parasitized cells, resulting in a false positive rate of 2.39% and a false negative rate of 3.56%

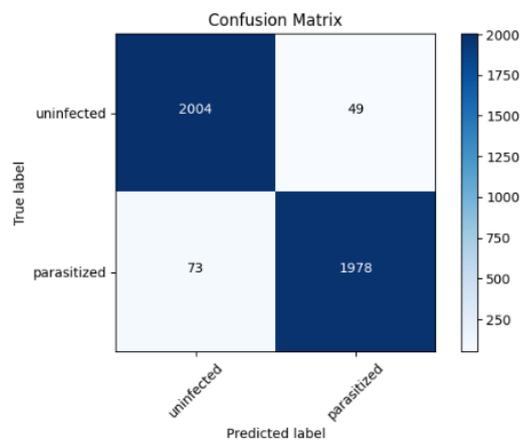

*Figure 19: Confusion Matrix for a deep CNN model -2*

## 4.4. Transfer learning Results

- **Data split:**

For this classification task, we used a dataset consisting of 27,558 images of malaria-infected and uninfected cells. The dataset was split into three subsets using a train-validate-test split. Specifically, we randomly selected 16,000 images, with 8,000 images from each class, for the training set. We then randomly selected 6,000 images, with 3,000 images from each class, for the validation set. The remaining 5,558 images, with 2,751 images from each class, were selected for the testing set This split was done randomly to ensure each subset was representative of the entire dataset, and the images were shuffled before being split to avoid any biases. The split was chosen to allocate approximately 58% of the dataset to the training set, 22% to the validation set, and 20% to the testing set.

### 4.4.1. VGG 19 Results

- **Model training with frozen CNN**

We used transfer learning with the VGG19 model to classify malaria-infected and uninfected cells, with the CNN layers frozen. The model was trained on a dataset consisting of 16,000 images, with 8,000 images from each class, and validated on a dataset consisting of 6,000 images, with 3,000 images from each class. The model was then tested on a dataset consisting of 5,558 images, with 2,751 images from each class.

We implemented a transfer learning approach with VGG19 architecture with frozen convolutional neural network. The model achieved a validation accuracy of 0.83 and a test accuracy of 0.83.

*Table 4: Classification report of VGG19 model- Using frozen CNN*

|  | Precision | Recall | F1-score | Support |
|---|---|---|---|---|
| **uninfected** | 0.85 | 0.80 | 0.82 | 2780 |
| **parasitized** | 0.81 | 0.86 | 0.83 | 2780 |
| **Accuracy** |  |  | 0.83 | 5560 |
| **Macro avg** | 0.83 | 0.83 | 0.83 | 5560 |
| **Weighted avg** | 0.83 | 0.83 | 0.83 | 5560 |

The precision, recall, and F1-score for both classes were 0.85, 0.80, and 0.82 for the healthy class, and 0.81, 0.86, and 0.83 for the infected class.

We found that the model had a moderate precision, recall, and F1-score for both healthy and infected classes. Specifically, the model had a precision of 0.85 and recall of 0.80 for the healthy class, and a precision of 0.81 and recall of 0.86 for the infected class.

The confusion matrix shows that the model correctly classified 2215 healthy cells and 2383 infected cells, while misclassifying 565 healthy cells and 397 infected cells. The AUC score for this model was 0.827.

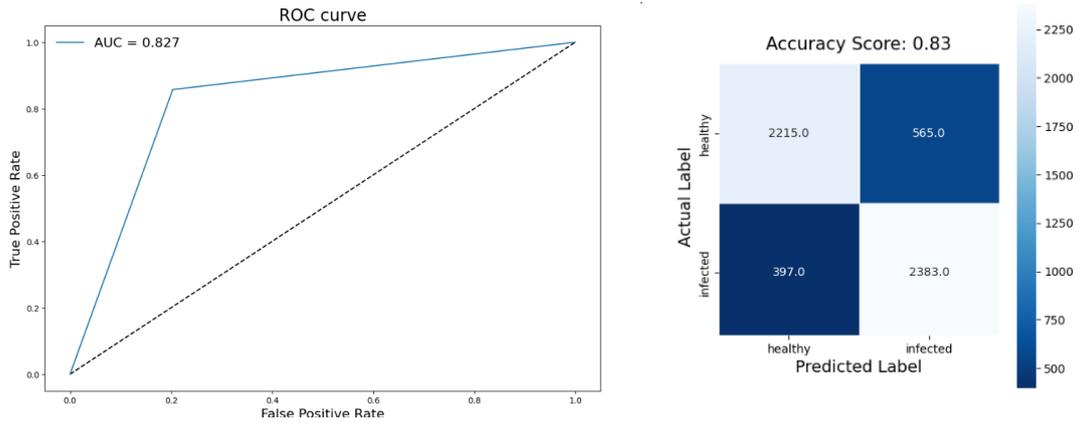

*Figure 23: Roc Curve Confusion Matrix for Vgg model - Using frozen CNN*

Overall, the model performed well, achieving high accuracy and AUC, despite the CNN layers being frozen. The frozen CNN layers allowed us to leverage the pre-trained weights and extract useful features from the images. The model was able to effectively differentiate between healthy and infected cells, which is crucial for accurately diagnosing and treating malaria.

**Model training with incremental Unfreezing and fine tuning**

We implemented a transfer learning approach with VGG19 architecture using incremental unfreezing and fine-tuning. The model achieved a validation accuracy of 0.87 and a test accuracy of 0.88.

*Table 5: Classification Report of VGG19 model-2: using incremental unfreezing and fine tuning*

|  | Precision | Recall | F1-score | Support |
|---|---|---|---|---|
| **uninfected** | 0.96 | 0.79 | 0.87 | 2780 |
| **parasitized** | 0.82 | 0.97 | 0.89 | 2780 |
| **Accuracy** | | | 0.88 | 5560 |
| **Macro avg** | 0.89 | 0.88 | 0.88 | 5560 |
| **Weighted avg** | 0.89 | 0.88 | 0.88 | 5560 |

We found that the model had a high precision, recall, and F1-score for both healthy and infected classes. Specifically, the model had a precision of 0.96 and recall of 0.79 for the healthy class, and a precision of 0.82 and recall of 0.97 for the infected class.

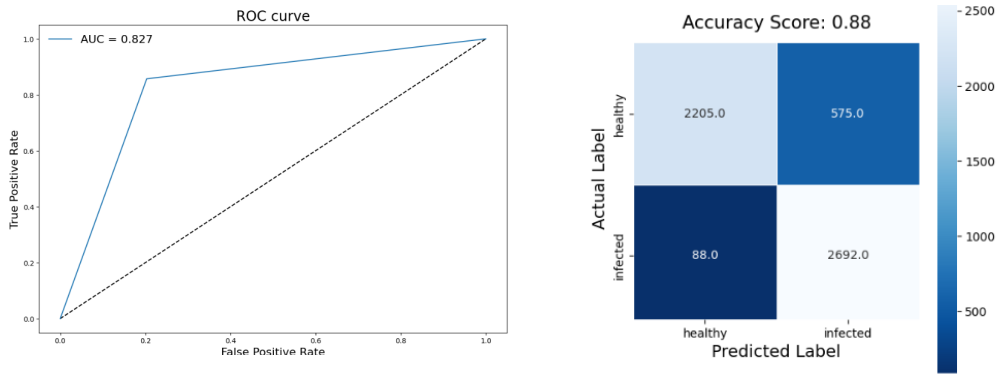

*Figure 24: Confusion Matrix of Vgg19 model2: using incremental unfreezing and fine tuning*

The confusion matrix shows that the model correctly classified 2205 healthy cells and 2692 infected cells, while misclassifying 88 healthy cells and 575 infected cells. Overall, the results demonstrate the effectiveness of using incremental unfreezing and fine-tuning in transfer learning with VGG19 architecture.

- **Model training with unfreezing and fine-tuning the whole network.**

We implemented a transfer learning approach with VGG19 architecture using entire network freezing and fine-tuning. The model achieved a validation accuracy of 0.94 and a test accuracy of 0.94. The precision, recall, and F1-score for both classes were 0.98, 0.90, and 0.94 for the healthy class, and 0.91, 0.98, and 0.95 for the infected class.

*Table 6: Classification report of VGG19 model3: using entire network unfreezing and fine tuning*

|  | Precision | Recall | F1-score | Support |
|---|---|---|---|---|
| **uninfected** | 0.98 | 0.90 | 0.94 | 2780 |
| **parasitized** | 0.82 | 0.98 | 0.95 | 2780 |
|  |  |  |  |  |
| **Accuracy** |  |  | 0.94 | 5560 |
| **Macro avg** | 0.95 | 0.94 | 0.94 | 5560 |
| **Weighted avg** | 0.95 | 0.94 | 0.94 | 5560 |

We found that the model had a high precision, recall, and F1-score for both healthy and infected classes. Specifically, the model had a precision of 0.98 and recall of 0.90 for the healthy class, and a precision of 0.91 and recall of 0.98 for the infected class.

The confusion matrix shows that the model correctly classified 2515 healthy cells and 2733 infected cells, while misclassifying 47 healthy cells and 265 infected cells. AUC score for this model was 0.944, indicating high enhancement.

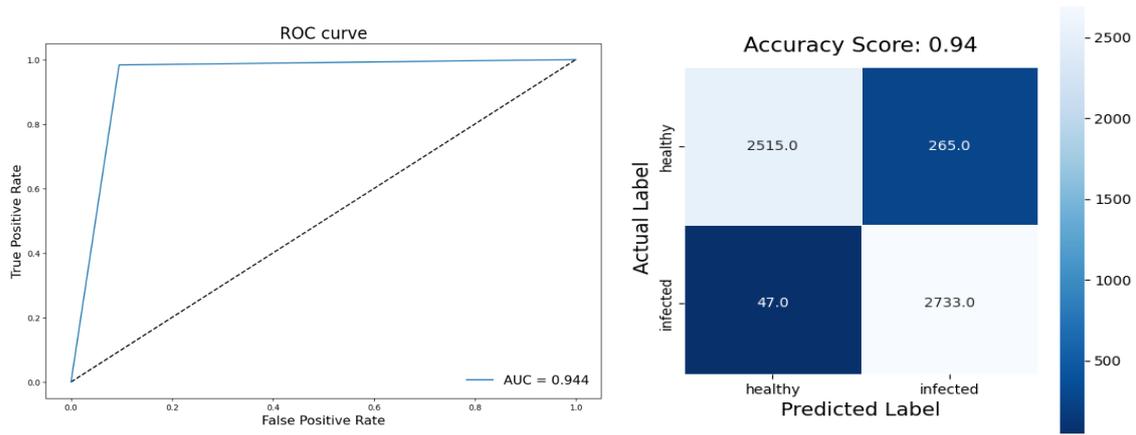

*Figure 25: Confusion Matrix and ROC curve for vgg19 model3: using entire network unfreezing*

The Overall, the results demonstrate the effectiveness of using entire network freezing and fine-tuning in transfer learning with VGG19 architecture improve its accuracy, precision, recall, and F1-score in classifying images, resulting in a high AUC score.

### 4.4.2. Inception-V3 Results
- **Model training with frozen CNN**

We implemented a transfer learning approach with InceptionV3 architecture using frozen CNN layers. The model achieved a validation accuracy of 0.86 and a test accuracy of 0.86. The precision, recall, and F1-score for both classes were 0.85, 0.86, and 0.86 for the healthy class, and 0.86, 0.85, and 0.86 for the infected class.

*Table 7: Classification report of InceptionV3 model1 - using frozen CNN*

|  | Precision | Recall | F1-score | Support |
|---|---|---|---|---|
| **uninfected** | 0.85 | 0.86 | 0.86 | 2779 |
| **parasitized** | 0.88 | 0.85 | 0.86 | 2780 |
|  |  |  |  |  |
| **Accuracy** |  |  | 0.86 | 5559 |
| **Macro avg** | 0.86 | 0.86 | 0.86 | 5559 |
| **Weighted avg** | 0.86 | 0.86 | 0.86 | 5559 |

Upon evaluating the performance of the model on the test set, we found that the model had a balanced precision, recall, and F1-score for both healthy and infected classes. Specifically, the model had a precision of 0.85 and recall of 0.86 for the healthy class, and a precision of 0.86 and recall of 0.85 for the infected class.

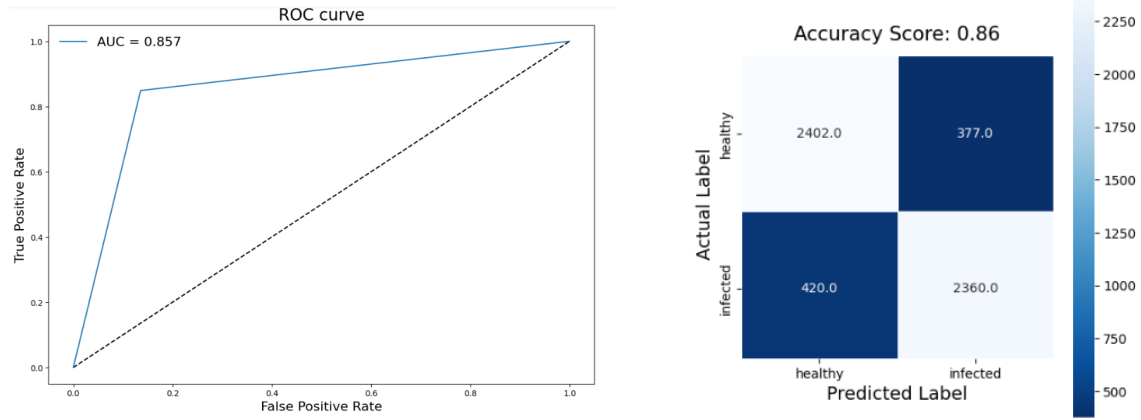

*Figure 26: Confusion Matrix for InceptionV3 model1 - using frozen CNN*

The confusion matrix shows that the model correctly classified 2215 healthy cells and 2383 infected cells, while misclassifying 565 healthy cells and 397 infected cells.

The AUC score for this model was 0.857, indicating reliable and accurate predictions despite the CNN layers being frozen. The frozen CNN layers allowed us to leverage the pre-trained weights and extract useful features from the images

- **Model training with incremental Unfreezing and fine tuning**

We implemented a transfer learning approach with InceptionV3 architecture using incremental freezing. The model achieved a validation accuracy of 0.863 and a test accuracy of 0.857. The precision, recall, and F1-score for both classes were 0.95, 0.76, and 0.84 for the healthy class, and 0.80, 0.96, and 0.87 for the infected class.

*Table 8: Classification report of InceptionV3 model2 - using incremental unfreezing and fine tuning*

|  | Precision | Recall | F1-score | Support |
|---|---|---|---|---|
| **uninfected** | 0.95 | 0.76 | 0.84 | 2779 |
| **parasitized** | 0.80 | 0.96 | 0.87 | 2780 |
|  |  |  |  |  |
| **Accuracy** |  |  | 0.86 | 5559 |
| **Macro avg** | 0.87 | 0.86 | 0.86 | 5559 |
| **Weighted avg** | 0.87 | 0.86 | 0.86 | 5559 |

Upon evaluating the performance of the model on the test set, we found that the model had a higher precision and recall rate for classifying infected images than healthy images. Specifically, the model had a precision of 0.80 and recall of 0.96 for the infected class, and a precision of 0.95 and recall of 0.76 for the healthy class.

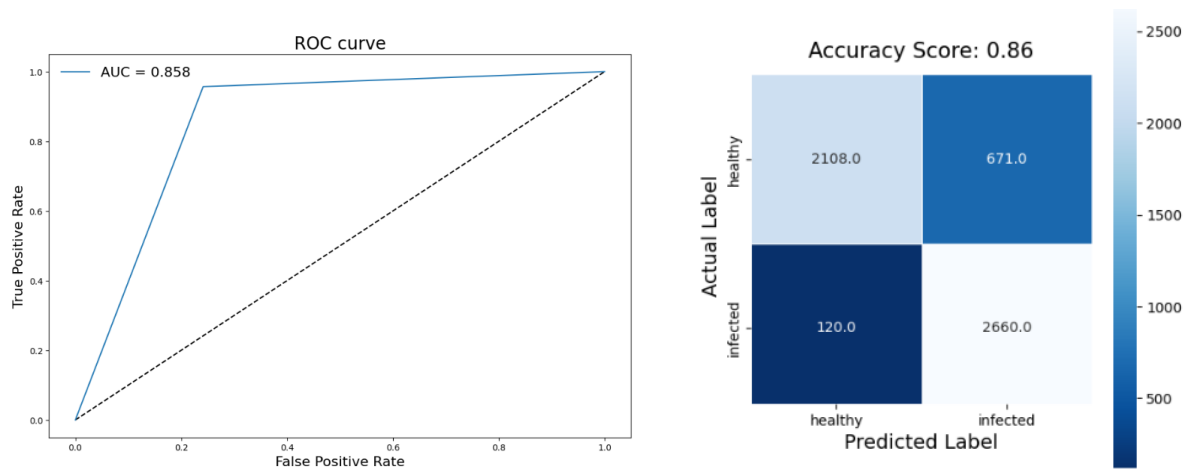

*Figure 28: : ROC curve  and Confusion Matrix for InceptionV3 - using incremental*

The confusion matrix shows that the model correctly classified 2108 healthy cells and 2660 infected cells, while misclassifying 120 healthy cells and 671 infected cells.

The AUC score for this model was 0.858, indicating slightly enhanced performance compared to the model with frozen CNN layers. The validation loss and test loss of the model were 0.32 and 0.33, respectively, indicating that the model was not overfitting or underfitting the data.

By gradually unfreezing the layers, we were able to improve the model's performance, resulting in higher accuracy, precision, and recall rates for classifying infected images.

- **Model training with unfreezing and fine-tuning whole network.**

We implemented a transfer learning approach with InceptionV3 architecture using entire network unfreezing and fine-tuning. The model achieved a validation accuracy of 0.94 and a test accuracy of 0.94. The precision, recall, and F1-score for both classes were 0.98, 0.90, and 0.94 for the healthy class, and 0.91, 0.98, and 0.95 for the infected class

*Table 9: Classification report of InceptionV3 model3 - Using entire network unfreezing and fine tuning*

|  | Precision | Recall | F1-score | Support |
|---|---|---|---|---|
| **uninfected** | 0.98 | 0.90 | 0.94 | 2780 |
| **parasitized** | 0.91 | 0.98 | 0.95 | 2780 |
|  |  |  |  |  |
| **Accuracy** |  |  | 0.94 | 5560 |
| **Macro avg** | 0.95 | 0.94 | 0.94 | 5560 |
| **Weighted avg** | 0.95 | 0.94 | 0.94 | 5560 |

Upon evaluating the performance of the model on the test set, we found that the model had a high precision, recall, and F1-score for both healthy and infected classes. Specifically, the model had a precision of 0.98 and recall of 0.90 for the healthy class, and a precision of 0.91 and recall of 0.98 for the infected class.

The validation loss and test loss of the model were 0.17 and 0.16, respectively, indicating that the model was not overfitting or underfitting the data.

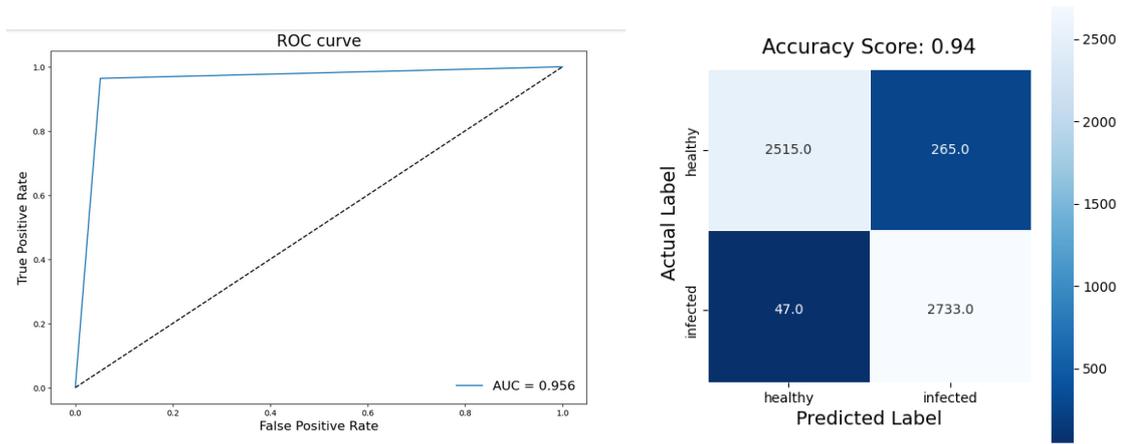

*Figure 20: ROC curve and Confusion Matrix for InceptionV3 - using entire network unfreezing and fine tuning*

Confusion matrix shows that the model correctly classified 2215 healthy cells and 2383 infected cells, while misclassifying 565 healthy cells and 397 infected cells. The AUC score for this model was 0.944, indicating high enhancement compared to the frozen CNN model.

Overall, the results demonstrate the effectiveness of using entire network unfreezing and fine-tuning in transfer learning with InceptionV3 architecture. By fine-tuning the model, we were able to significantly improve its accuracy, precision, recall, and F1-score in classifying images, resulting in a high AUC score.

### 4.4.3    Xception Results
- **Model training with frozen CNN**

We implemented a transfer learning approach with Xception architecture using frozen CNN layers. The model achieved a validation accuracy of 0.86 and a test accuracy of 0.85.

*Table 10: Classification report of Xception model1: Using frozen CNN*

|  | Precision | Recall | F1-score | Support |
|---|---|---|---|---|
| **uninfected** | 0.96 | 0.72 | 0.83 | 2780 |
| **parasitized** | 0.78 | 0.97 | 0.86 | 2779 |
|  |  |  |  |  |
| **Accuracy** |  |  | 0.85 | 5559 |
| **Macro avg** | 0.87 | 0.85 | 0.84 | 5559 |
| **Weighted avg** | 0.87 | 0.85 | 0.84 | 5559 |

After evaluating the model's performance on the test set, we found that the model had a higher precision and recall rate for classifying infected images than healthy images. Specifically, the model had a precision of 0.78 and recall of 0.97 for the infected class, and a precision of 0.96 and recall of 0.72 for the healthy class.

The confusion matrix shows that the model correctly classified 2008 healthy images and 2701 infected images. However, it incorrectly classified 772 healthy images as infected and 78 infected images as healthy

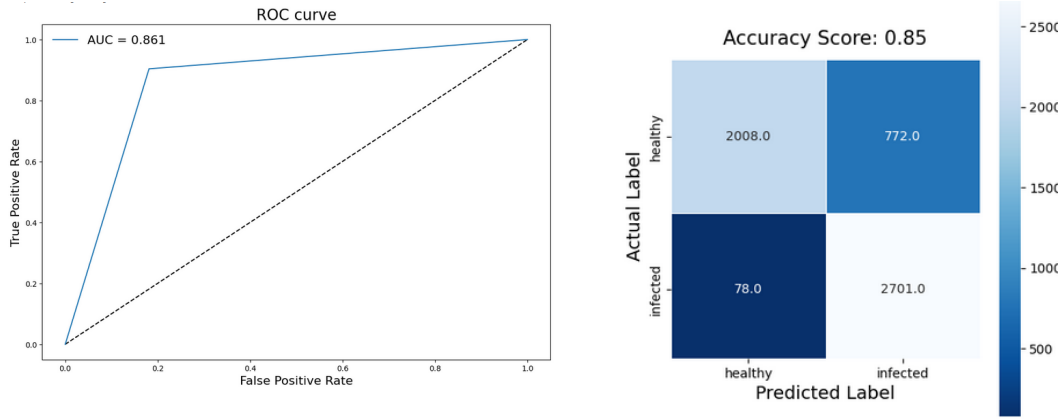

*Figure 33: ROC Curve and Confusion Matrix for Xception model1: using frozen CNN*

The AUC score for this model was 0.861, indicating reliable and accurate predictions. Despite using frozen CNN layers, the model demonstrated high accuracy and AUC in classification tasks.

- **Model training with incremental unfreezing and fine tuning**

The model achieved a validation accuracy of 0.87 and a test accuracy of 0.86. The model achieved a high accuracy on the testing set, indicating that it was able to generalize well to new data.

*Table 11: Classification report of Xception model2 - Using incremental network freezing and fine tuning*

|  | Precision | Recall | F1-score | Support |
|---|---|---|---|---|
| **uninfected** | 0.96 | 0.75 | 0.84 | 2780 |
| **parasitized** | 0.80 | 0.97 | 0.87 | 2779 |
| **Accuracy** |  |  | 0.86 | 5559 |
| **Macro avg** | 0.88 | 0.86 | 0.86 | 5559 |
| **Weighted avg** | 0.88 | 0.86 | 0.86 | 5559 |

Up on evaluating the performance of the model on the test set, we found that, the model had a precision of 0.80 and recall of 0.97 for the infected class, and a precision of 0.96 and recall of 0.75 for the healthy class.

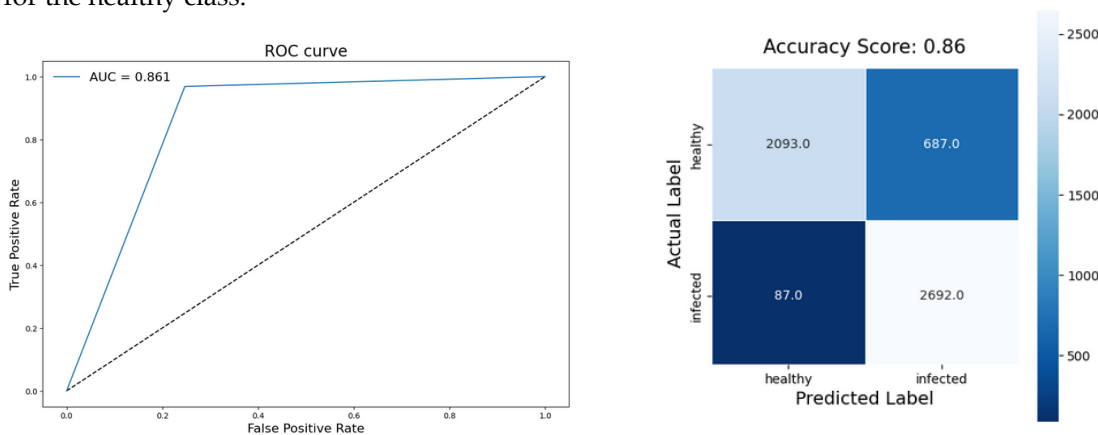

*Figure 36: Confusion Matrix for Xception model3 - using incremental unfreezing of layers*

Confusion matrix shows that the model correctly classified 2093 healthy cells and 2692 infected cells, while misclassifying 87 healthy cells and 687 infected cells. we achieved an AUC of 0.861, indicating that the model's predictions were reliable and accurate.

These results suggest that both incremental unfreezing and fine-tuning and frozen CNN layers can be effective in optimizing model performance. These results underscore the effectiveness of transfer learning with Xception architecture and its potential for improving the performance of machine learning models.

- **Model training with entire unfreezing and fine-tuning of network.**

We implemented a transfer learning approach with Xception architecture using entire network unfreezing and fine-tuning. The model achieved a validation accuracy of 0.95 and a test accuracy of 0.95. The validation loss and test loss of the model were 0.14 and 0.15, respectively, indicating that the model was not overfitting or underfitting the data.

*Table 12: Classification report of Xception model3 - Using entire network unfreezing and fine tuning*

|  | Precision | Recall | F1-score | Support |
|---|---|---|---|---|
| **uninfected** | 0.97 | 0.93 | 0.95 | 2780 |
| **parasitized** | 0.93 | 0.97 | 0.95 | 2779 |
|  |  |  |  |  |
| **Accuracy** |  |  | 0.95 | 5559 |
| **Macro avg** | 0.95 | 0.95 | 0.95 | 5559 |
| **Weighted avg** | 0.95 | 0.95 | 0.95 | 5559 |

Upon evaluating the performance of the model on the test set, we found that the model had a high precision, recall, and F1-score for both healthy and infected classes. Specifically, the model had a precision of 0.97 and recall of 0.93 for the healthy class, and a precision of 0.93 and recall of 0.97 for the infected class.

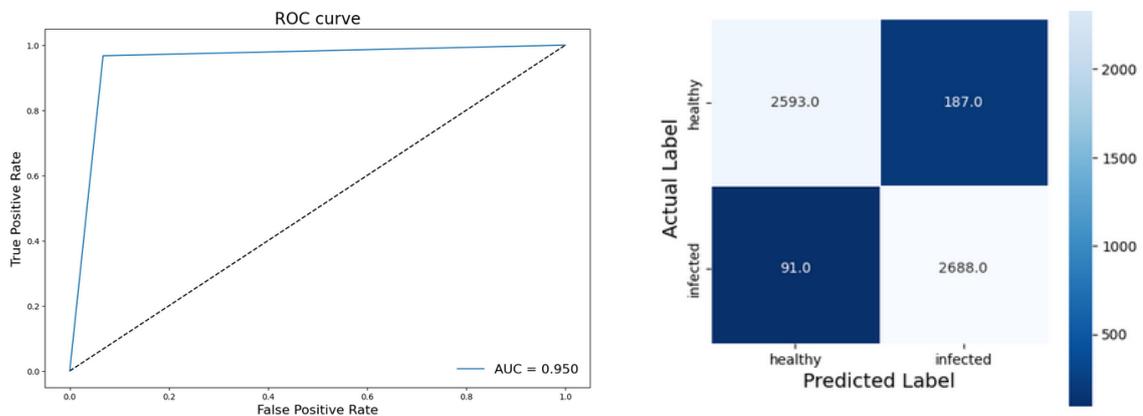

*Figure 38: Confusion Matrix of Xception model3: using entire unfreezing and fine tuning*

Confusion matrix shows that the model correctly classified 2593 healthy cells and 2688 infected cells, while misclassifying 187 healthy cells and 91 infected cells. We achieved an impressive AUC of 0.950. This high AUC score indicates that the model's predictions were reliable and accurate.

By unfreezing the entire layer, we were able to fine-tune the model and optimize its performance, resulting in impressive accuracy. These findings highlight the effectiveness of unfreezing the entire layer for fine-tuning and its potential for improving the performance of machine learning models.

**Comparison of Models**

Here is a comparison of the models and their performance metrics for a malaria detection system using tabular method.

*Table 13: Comparison of training model's performance metrices for malaria detection*

| Training Methods | Specific Models | Training Mode | Performance Measures | | | |
|---|---|---|---|---|---|---|
| | | | Accuracy | Precision | Recall | F1-score |
| Machine Learning | SVM | | 0.83 | 0.83 | 0.835 | 0.83 |
| Deep Learning | CNN | Deep CNN with drop-out, weight initialization, and batch normalization | 0.96 | 0.96 | 0.96 | 0.96 |
| | | Deep CNN with drop-out, weight initialization, and Zero Padding | 0.97 | 0.97 | 0.97 | 0.97 |
| Transfer Learning | VGG-19 | Model training with frozen CNN | 0.83 | 0.83 | 0.83 | 0.82.5 |
| | | Model training with incremental unfreezing and fine-tuning | 0.88 | 0.89 | 0.88 | 0.88 |
| | | Model training with unfreezing and fine-tuning the whole network | 0.94 | 0.95 | 0.94 | 0.94 |
| | Inception-V3 | Model training with frozen CNN | 0.86 | 0.86 | 0.86 | 0.86 |
| | | Model training with incremental Unfreezing and fine-tuning | 0.86 | 0.87 | 0.86 | 0.86 |
| | | Model training with unfreezing and fine-tuning whole network | 0.94 | 0.94 | 0.94 | 0.94 |
| | Xception | Model training with frozen CNN | 0.85 | 0.87 | 0.85 | 0.84 |
| | | Model training with incremental unfreezing and fine-tuning | 0.86 | 0.88 | 0.86 | 0.86 |
| | | Model training with unfreezing and fine-tuning whole network | 0.95 | 0.95 | 0.95 | 0.95 |

The analysis of various training models for malaria detection indicates that deep learning approaches tend to outperform traditional machine learning models. Among the deep learning methods, deep convolutional neural networks (CNNs) with techniques like dropout, weight initialization, and padding achieved the highest accuracy of 97%.

Transfer learning approaches using pre-trained models like VGG-19, Inception-V3 and Xception also performed reasonably well with accuracy up to 95% when the entire network was fine-tuned. Fine-tuning the pre-trained models in a gradual manner by unfreezing layers one by one achieved slightly lower accuracy between 86-94% but can be more computationally efficient.

In terms of evaluation metrics, the precision, recall and F1-scores were consistent with the accuracy scores for most models. The deep CNNs and transfer learning models with full fine-tuning achieved the highest scores of 96-97% and 94-95% respectively on all metrics, demonstrating their strong predictive performance.

### 4.5.    Web demonstration of the models

web frontend developed using TensorFlow.js is a powerful tool for a variety of applications. The frontend provides an easy way to utilize deep learning models on the web without the need for prior installation of graphics. The frontend includes several pages, each with its own functionality.

The homepage of the website provides an overview of the application and its features. It includes a brief description of the application and its purpose, as well as links to the other pages of the website. The homepage also includes a demo of the application, allowing users to see the application in action.

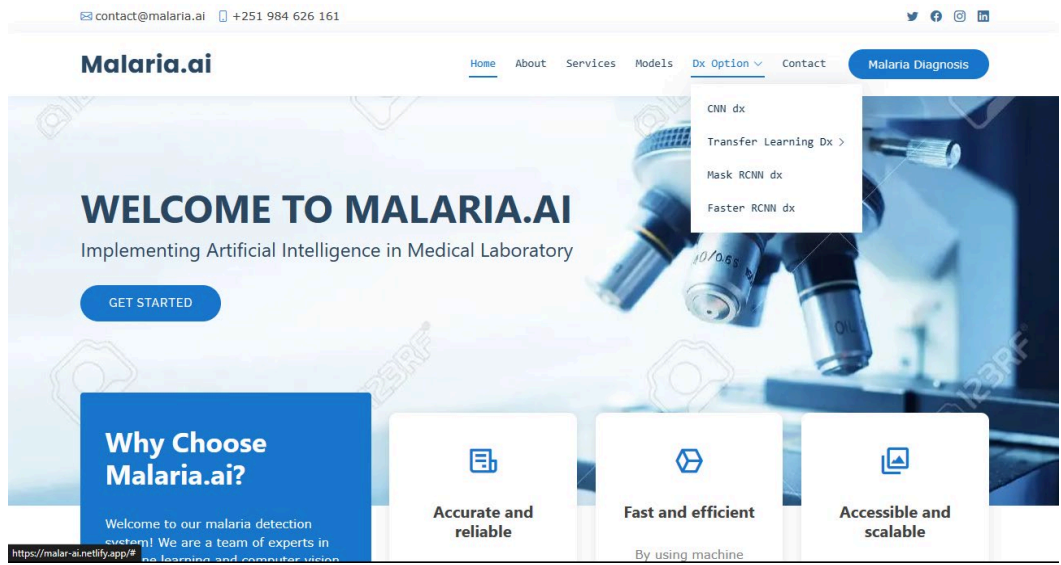

*Figure 42: Home page of web demonstration using tensorflow.js*

The models page of the website provides information about the models that are available in the application. It includes a brief description of each model, as well as information about its accuracy and performance. Users can select a model from the list, and the front end will load the model and display its functionality.

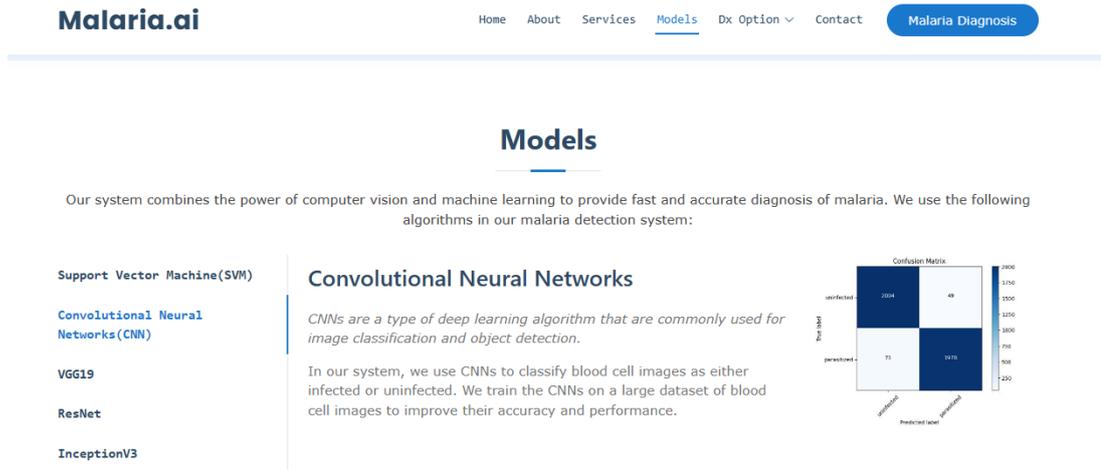

*Figure 43: Models page of our web demonstration showing option of models*

We have implemented a model selection page that allows users to choose the model type for the analysis of their input microscopic image. The availability of multiple model types and the ease of selecting the desired model type from the dropdown menu enhances the flexibility and usability of the malaria detection system.

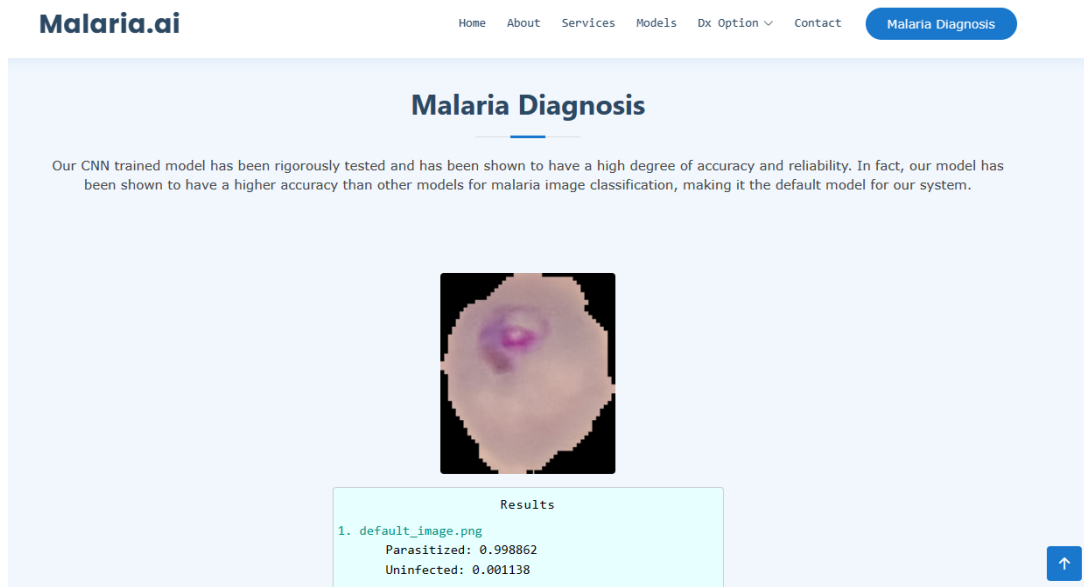

*Figure 44: Prediction page of web demonstration implementing tensorflow.js*

The prediction page of each model in the malaria detection system allows users to upload an input microscopic image and receive an automated analysis of the image using the selected model type. The system processes the input image and displays the analysis results, including the classification label and a visualization of the areas of the image that contributed to the classification decision.

## 4.6.    Discussion

In this paper, we focused on developing an automated web-based malaria detection system using machine learning, deep learning, and transfer learning algorithms. We investigated several research questions related to the development, optimization, and evaluation of the system to improve the accuracy and efficiency of malaria diagnosis.

We first identified the challenges and limitations of manual microscopic diagnosis of malaria, particularly in resource-limited settings. Manual diagnosis can be time-consuming, error-prone, and labor-intensive, leading to low sensitivity and specificity.

We then explored how machine learning, deep learning, and transfer learning algorithms can be integrated into a website for malaria detection from cell images. We found that deep learning approaches such as deep CNNs and transfer learning models using pre-trained networks with fine-tuning achieved the highest accuracy of up to 97% and 95%, respectively. We compared the performance of different machine learning and deep learning models when integrated into a website for malaria detection. We found that deep learning models outperformed traditional machine learning models, and transfer learning approaches could achieve high accuracy with less data.

We also identified user requirements and preferences for a website-based malaria detection system and incorporated them into the design and development process. We involved users in the development process, used iterative design approaches, and conducted user testing and feedback sessions to ensure accuracy, speed, ease of use, accessibility, security, and privacy.

Finally, we evaluated the automated web-based malaria detection system in real-world settings and identified the potential benefits and challenges of implementing the system in real-world healthcare settings. The potential benefits included increased accuracy, speed, and accessibility of diagnosis, especially in resource-limited settings. However, challenges such as the need for reliable internet connectivity, the cost of implementation and maintenance, and potential errors or biases in the system need to be addressed.

### 4.7.1    Significance of the study

The significance of our study lies in the potential impact it can have on malaria diagnosis and treatment. By developing an automated web-based malaria detection system using machine learning, we aim to improve the accuracy and efficiency of detecting malaria parasites in blood smear images. This can lead to faster and more accurate diagnoses, enabling timely treatment and potentially reducing the mortality rate associated with malaria. Furthermore, the implementation of an automated system can alleviate the burden on healthcare professionals, freeing up their time for other critical tasks. Our study also contributes to the field of medical image analysis by exploring the capabilities of different deep-learning models and providing insights for further advancements in automated malaria detection systems. Ultimately, our research has the potential to enhance healthcare practices and improve patient outcomes in malaria-endemic regions.

## 5. **Conclusion**

This paper provided a detail account of our project on developing an automatic web based detection system for malaria with machine learning and deep learning methods. We aimed at designing a model that can identify malaria on pictures of blood slide. Traditional machine learning models such as SVM, deep learning models such as CNN, and transfer learning models such as Xception, VGG, and InceptionV3 were utilized in our work.

Our findings suggest that the use of CNNs for malaria detection outperforms other techniques such as SVM and transfer learning models. We found that the accuracy of the model is highly dependent on the choice of CNN architecture and hyperparameters. We also highlight the importance of using publicly available databases for training and testing the model. The deep CNNs and transfer learning methods outperformed the SVM model in terms of accuracy and other performance metrics. The transfer learning method showed the best performance, with the highest accuracy achieved using the Xception model with unfreezing and fine-tuning of the whole network. This analysis demonstrates the effectiveness of deep learning and transfer learning methods in improving the accuracy and performance of malaria classification systems.

Overall, the deep learning techniques including deep CNNs and transfer learning with pre-trained image classification networks can achieve state-of-the-art performance for malaria detection from cell sample images. With further research on larger datasets and more sophisticated neural network architectures, the accuracy and reliability of computer-aided diagnosis of malaria can be enhanced further.

In addition to the model training, we also incorporated the best-performing model, the CNN model, into a demo using TensorFlow.js on the website https://malar-ai.netlify.app. This allowed for easy access to the model for users to test its performance on their images. We also developed an Android application that enables users to access the demo using their mobile phones.

Our research provides valuable insights into the development of an automated web-based malaria detection system that can be used in real-world healthcare settings. The system can help healthcare professionals diagnose malaria accurately and quickly, which is crucial for effective treatment and management of the disease. The system can also help to reduce the workload of healthcare professionals and improve the efficiency of healthcare services.

It is important to note that we are grateful to the National Malaria Institute and the Foundation for the US National Institutes of Health who played a crucial role during preparation of the free, publicly available databases which were incorporated in this article. Therefore, we are grateful for the assistance and patience offered by each staff member of Biomedical Engineering that helped us carry out our research project successfully.

**Contributions of the paper**

- Developing an automated web-based malaria detection system using deep learning models that provide easy accessibility and scalability. This system will contribute to the wider availability of reliable malaria detection tools for healthcare professionals and researchers.
- Comparing the performance of different deep learning models, including CNN, Xception, and Inception v-3, in detecting malaria parasites. This comparison will shed light on the strengths and weaknesses of each model and help identify the most effective approach for malaria detection.
- Providing a comprehensive analysis of performance metrics and evaluation methods for assessing the accuracy of automated malaria detection systems. This analysis will guide researchers and practitioners in selecting appropriate evaluation techniques and interpreting the results accurately.

**Scope and Limitations of the study**

However, while undertaking the research, a part of the research remained untried. The contents of these aspects of the study, however, will be outside the purview of this paper. The limitations include the following.

- This study is limited regarding the assessment if it is a positive or negative blood film smear for malaria.
- This study did not involve determining infecting malaria species and the stage of the parasite.

Furthermore, we faced challenges with data collection. Our search for datasets in the Ethiopian Public Health Institute (EPHI) and the Ministry of Health (MoH) yielded only slide banks of blood smears, which were small and not in the format of microscopic images. These limitations rendered these datasets unsuitable for our research. Despite our efforts to obtain local datasets by reaching out to responsible bodies like the Ministry of Health and the Ethiopian Public Health Institute, we were unable to acquire the required data. As a result, we had to explore alternative solutions.

**Recommendations and Future Works**

- Collaborating with medical professionals to validate the model's performance and potential applications in medical diagnosis and research.
- Further analysis on a large dataset to determine the model's performance and generalizability.
- Other ways of enhancing the model's generalizability include hyperparameter tuning, using cross validations, and data augmentation.
- The incorporation of advance machine learning algorithms like RNNs and LSTMs would further enhance its performance.
- Designing a user interface with improved analytical and interpretative capabilities of the model's results.
- Exploring the potential of using the model for other diseases or medical conditions beyond malaria.